\documentclass[runningheads]{llncs}
\usepackage[utf8]{inputenc} %
\usepackage[T1]{fontenc}    %
\usepackage{url}            %
\usepackage{booktabs}       %
\usepackage{amsmath}
\usepackage{amsfonts}       %
\usepackage{microtype}      %
\usepackage[pdftex]{graphicx} %
\graphicspath{{./figures/}}
\usepackage{pgfplots}
\pgfplotsset{compat=1.13} 
\usepackage{tikz}
\usetikzlibrary{shapes.geometric, arrows}
\usepackage{xcolor}
\usepackage{placeins}
\usepackage[misc,geometry]{ifsym}

\newcommand{\etal}{{\em et al.~}}

\begin{document}
\title{Real-Time 2D-3D Deformable Registration with Deep Learning and Application to Lung Radiotherapy Targeting}
\titlerunning{Real-Time 2D-3D Deformable Registration with Deep Learning}
\author{Markus D. Foote\inst{1}$^{\text{\Letter}}$
	\orcidID{0000-0002-5170-1937}\thanks{The final authenticated publication is available online at \texttt{https://doi.org/10.1007/978-3-030-20351-1\_20}.} \and %
Blake E. Zimmerman\inst{1}\orcidID{0000-0003-1769-7943} \and
Amit Sawant\inst{2} \and %
Sarang C. Joshi\inst{1}%
}
\authorrunning{M. Foote et al.}
\institute{Scientific Computing and Imaging Institute, Department of Bioengineering, University of Utah, Salt Lake City, UT, USA\\ \email{foote@sci.utah.edu} \and
Department of Radiation Oncology, University of Maryland, School of Medicine, Baltimore, MD, USA }
\maketitle              %
\begin{abstract}
Radiation therapy presents a need for dynamic tracking of a target tumor volume.
Fiducial markers such as implanted gold seeds have been used to gate radiation delivery but the markers are invasive and gating significantly increases treatment time.
Pretreatment acquisition of a respiratory correlated 4DCT allows for determination of accurate motion tracking which is useful in treatment planning.
We design a patient-specific motion subspace and a deep convolutional neural network to recover anatomical positions from a single fluoroscopic projection in real-time.
We use this deep network to approximate the nonlinear inverse of a diffeomorphic deformation composed with radiographic projection.
This network recovers subspace coordinates to define the patient-specific deformation of the lungs from a baseline anatomic position. 
The geometric accuracy of the subspace deformations on real patient data is similar to accuracy attained by original image registration between individual respiratory-phase image volumes.
\end{abstract}
\section{Introduction}

 According to the United States Centers for Disease Control and Prevention, lung cancer is the leading cause of cancer death, accounting for 27\% of all cancer deaths in the United States \cite{cdc}.
 Accurate estimation of organ movement and normal tissue deformations plays a crucial role in dose calculations and treatment decisions in radiation therapy of lung cancer \cite{Geneser2011}.
State-of-the-art radiation treatment planning uses 4D (3D + time) respiratory correlated computed tomography (4D-RCCT) scans as a planning tool to deliver radiation dose to the tumor during treatment \cite{Keall2005} and minimize dose to the healthy tissue and vital organs. Understanding respiratory motion allows for more accurate treatment planning and delivery, resulting in more targeted dose delivery to only the tumor.
 
 While general motion patterns in radiotherapy patients are relatively well understood, cycle-to-cycle variations remain a significant challenge and may account for observed discrepancies between predictive treatment plan indicators and clinical patient outcomes, especially due to an increased irradiated volume which limits adequate dose to the target \cite{Sawant2014}.
 Gating methods have been developed but require implanted markers and lengthen treatment time.
 Real-time motion tracking using magnetic resonance imaging is only applied within the 2D imaging slices due to time constraints for imaging \cite{Ha2019a}. 
 Accurate and fast noninvasive target tracking that accounts for real-time cycle-to-cycle variations in respiratory motion is a recognized need in radiotherapy \cite{Markelj2012,Sawant2008}.
 
 In this paper, we propose to use 4D-RCCT acquired at treatment planning time to develop a finite low dimensional patient specific space of deformations and use treatment time single angle fluoroscopy imaging in conjunction with deep convolutional neural network to recover the full deformations of the anatomy during treatment.
 The constant time inference capability of convolutional neural networks allows real-time recovery of the anatomic position of the tumor target and surrounding anatomy using radiographic images of the treatment area acquired using fluoroscopy. 
We envision our framework to be incorporated as the target position monitoring subsystem in a conformal radiotherapy system \cite{Sawant2008}. 
 The proposed method eliminates the need for invasive fiducial markers while still producing targeted radiation delivery during variable respiration patterns.
 While this framework requires training of the model on a per-patient basis, only inference is required during treatment.
 Inference of the deep convolutional network and subsequent linear combination of patient-specific motion components can be calculated faster than real-time and drive motion-tracking of conformal radiotherapy treatment.
 
 The proposed framework begins by creating motion subspace from CT registration among the respiratory phase images. 
 This motion subspace is then used to generate many respiratory phase fluoroscopic images with known subspace coordinates.
 These labeled fluoroscopic images serve as a training dataset for a deep convolutional neural network that recovers the motion subspace coordinates.
 Motion coordinates recovered by the deep neural network from unseen fluoroscopy define a linear combination of deformation components that form a full 3D deformation that is represented in the 2D fluoroscopic image.

\section{Related Work}
\subsection{Low-rank Respiratory Motion Models}
The respiratory cycle exhibits low-rank behavior and hysteresis, leading to application of principal component analysis to study the deformation of lungs and other organs through time \cite{Li20116009}. 
Sabouri \etal ~2017 has used such a PCA approximations of the respiratory cycle to correlate the lung's principal deformations to respiratory markers directly observable during treatment such as body surface tracking during conformal radiation therapy \cite{Sabouri2017}.
PCA has also been applied to MR-based image guidance in radiotherapy \cite{Ha2019a}.

\subsection{Rank Constrained Density Motion Estimation}
\label{sec:rankdef}
Our real-time tracking procedure is based on the rank-constrained diffeomorphic density matching framework, summarized here for completeness (see~\cite{Bauer2015,Foote2017,Rottman2015density,Rottman2016} for details).
In general, the rank-constrained diffeomorphic density matching problem produces a registration between multiple image volumes that is explained by only a few principal components of deformation -- less than the total number of volumes. 
The deformation set matrix rank is non-smooth and therefore unfit for optimization. 
Instead, the nuclear norm is used as a smooth and convex surrogate. %
The deformation set matrix is constructed with each vectorized deformation field as a row, 
\begin{equation}
X=\left[\begin{matrix}
\varphi_1^{-1}(x) -x\\
\varphi_2^{-1}(x) -x\\
\vdots\\
\varphi_{N-1}^{-1}(x) - x
\end{matrix} \right]= \{\varphi_i^{-1}(x) - x\}
\end{equation}
where $\varphi_i^{-1}$ is the inverse of the deformation from the $i$-th image in the image series to a selected reference image.
We  thus define the nuclear norm for deformations between $N$ images as 
\begin{equation}
\left\| X \right\|_* = \sum_{i}^{N-1} |\sigma_i\left(X\right)| \ ,
\end{equation}
where $\sigma_i$ are the singular values.
As there are $N-1$ deformations between $N$ images, giving only $N-1$ singular values $\sigma$. 

As CT measures the linear attenuation coefficient which is proportional to the true density, 4DCT volumes are interpreted as densities, thus values change with compression or expansion. 
A density $I\, dx$ is acted upon by a diffeomorphism $\varphi$ to compensate for changes of the density by the deformation:
\begin{equation}
\left(\varphi, I\, dx\right)\mapsto \varphi_*\left(I\, dx\right) = \left(\varphi^{-1}\right)^*\left(I\, dx\right)=\left(|D\varphi^{-1}|I\circ\varphi^{-1}\right)dx
\end{equation}
where $|D\varphi^{-1}|$ denotes the Jacobian determinant of $\varphi^{-1}$.  
The Riemannian geometry of the group of diffeomorphisms with a suitable Sobolev $H^1$ metric is linked to the Riemannian geometry of densities with the Fisher-Rao metric \cite{Bauer2015,Khesin2013,Modin2015}. 
The Fisher-Rao metric is used as the measure for similarity due to the property that it is invariant to the action of diffeomorphisms:
\begin{equation}
d^2_F\left(I_0\, dx, I_1\, dx\right) = \int_{\Omega}\left(\sqrt{I_0}-\sqrt{I_1}\right)^2 dx \;.
\end{equation}

The linkage between a suitable Sobolev $H^1$ metric and the Fisher-Rao metric allows for evaluation of the distance in the space of diffeomorphisms in closed form. 
The Fisher-Rao metric, an incompressibility measure, and surrogate rank measure can then be used to match a set of densities by minimizing the energy functional:
\begin{multline}
E(\left\{\varphi_i\right\}) = \sum_{i}^{N-1}\left[ \int_{\Omega}\left(\sqrt{\left|D\varphi_i^{-1}\right|I_i\circ\varphi_i^{-1}}- \sqrt{I_0} \right)^2 dx \;  \right. \\
\left. +\int_{\Omega} \left(\sqrt{\left|D\varphi_i^{-1}\right|}-1\right)^2 f \,dx \right]  + \alpha \sum_{i}^{N-1} |\sigma_i\left(\{\varphi_i^{-1}(x) - x\}\right)|
\label{eq:energy}
\end{multline}
where $I_0$ is a chosen base or reference density and $I_i$ are the other $N-1$ densities in the series.
The first term here penalizes dissimilarity between the two densities. 
The second term penalizes deviations from a volume-preserving deformation. 
The penalty function $f$ acts as weighting of the volume-preserving measure. 
The final term penalizes the surrogate rank of the subspace in which the set of deformations exist. 
We use a gradient flow and and iterative soft thresholding for optimizing the above energy functional.
This optimization produces a related set of rank-optimized deformations with improved geometric accuracy over other registration methods due to the increased physiologic relevance of the low-rank deformations which match well with the general forward-reverse process of the inhale-exhale cycle, along with hysteresis in other components \cite{Foote2017}.

\subsection{Digitally Reconstructed Radiographs}
For a given 4D-RCCT dataset, it is sufficient for training to simulate the 2D projection images at different respiratory states from the 3D volumes acquired at different states. 
These digitally reconstructed radiographs (DRR) are commonly used for radiation therapy planning.
This allows for simulation of the 2D projections that would be acquired at the different phases of breathing throughout a state-of-the-art radiotherapy treatment \cite{Rottman2015,Sawant2008}.
A point $(u,v) \in \mathbb{R}^2$ in the DRR projection image is defined by the line integral of linear attenuation coefficients (values in the CT volume) over distance $l$ from the x-ray source to the projection image \cite{Markelj2012,Sherouse1990,Staub2013}. 
The path can be written as $p(s) = \left( s(u - u_0), s(v - v_0), sl\right)$ where $s \in [0,1]$ and $(u_0, v_0)$ is the piercing point, or the unique point in the projection image that is closest to the x-ray source.
The DRR operation generates a 2D projection of the 3D CT volume that can be used for training in place of actual radiography recordings. 
These projections closely approximate the fluoroscopic images acquired during a state-of-the-art radiotherapy treatment session.

\section{Methodology}

\subsection{Motion Subspace Learning} %
A 10-phase respiratory correlated 4D-RCCT of a lung cancer patient for treatment planning from the University of Maryland was provided in anonymized form. 
Each 3D volume in the 4D-RCCT dataset is $512\times512\times104$ with resolution $0.976\times0.976\times3$ mm.
These 10 3D CT volumes, each representing a distinct respiratory phase, are used for rank constrained density registration as described by Foote \etal~\cite{Foote2017} to produce 9 rank-optimized deformation fields that map to the full-exhale phase volume from each of the other phase volumes .
The low-rank subspace of the space of diffeomorphisms in which the rank-optimized deformation components exist is determined via principal component analysis (PCA).

\subsection{Training Dataset Generation}
\label{sec:synth}
\begin{figure}[ht]
	\centering
	\includegraphics[width=\textwidth]{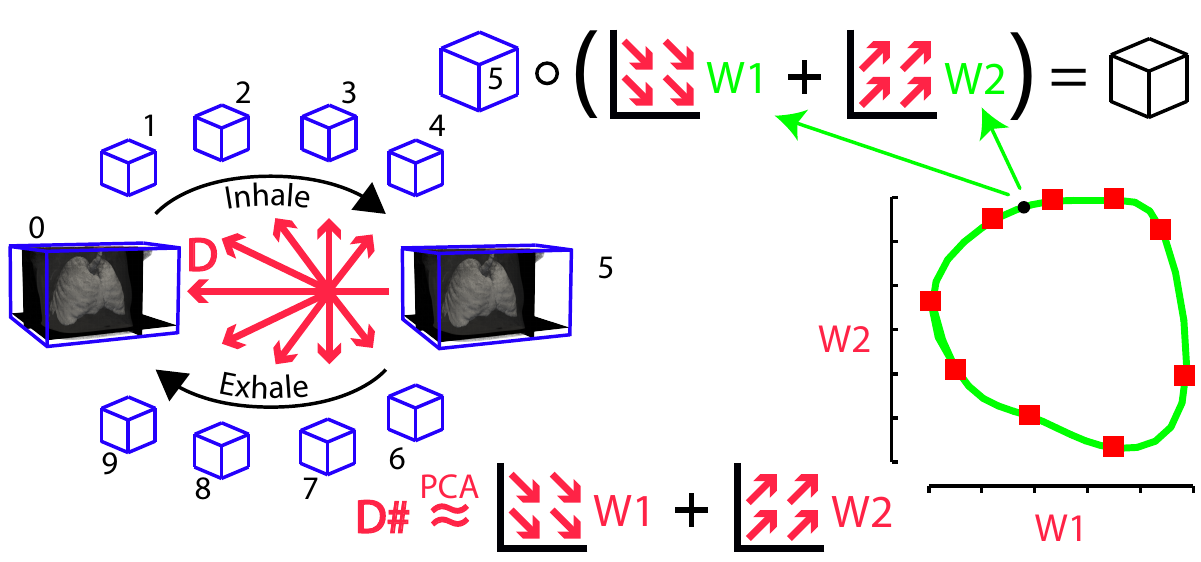}
	\caption{\label{fig:overview} Dataset generation overview. The deformations from rank constrained motion estimation are decomposed using PCA. Coordinates within the PCA space are used to generate a new deformation field, which is then applied to the reference CT volume. The resulting volume is used for DRR projection.}
\end{figure}

A dataset for training is generated by calculating DRR projections through a deformed reference CT volume %
The deformation applied to the reference volume is calculated as a linear combination of rank-optimized components.
Only the first two rank-optimized components are used as they explain 99.7\% of the deformation data (Table \ref{tab:pca}). 
The weights for this linear combination are generated as a dense grid of 300 evenly spaced points in first rank-optimized component direction and 200 evenly spaced points in the second (60000 total samples for training).
The value of the maximum weight is set as 150\% and 110\% of the maximum magnitude of the corresponding rank-optimized component for the first and second directions, respectively.
These weights serve as the target  for training and are then used to calculate a linear combination of the rank-optimized component deformations.
The resulting grid of deformations are calculated and applied to the full-exhale 4D-RCCT volume using PyCA \cite{pyca} (Fig. \ref{fig:overview}).
DRR images are calculated by ray casting and integrating along a cone-beam geometry of the fluoroscope through the deformed full-exhale 4DCT volume \cite{Rottman2015,Sherouse1990,Staub2013}.
The geometry setup for the DRR generation is similar to the geometry that would be found in a current radiotherapy treatment linear accelerator's imaging accessories, such as the Varian TrueBeam. 
Specifically, the x-ray source was positioned laterally from the center of the CT volume.
The center of the virtual detector for the projection image was then positioned 150 cm  horizontally from the source, through the CT volume, with a $1024\times768$ pixel array with spatial resolution $0.388\times0.388\:$mm. %
DRR images were preprocessed before training with variance equalization, normalizing the image intensities to $[0,1]$, histogram equalization, and $2\times2$ average downsampling.

\begin{table}[htb]
	\caption{Data explanation power of cumulative rank-optimized components of deformation from rank-constrained density deformation optimization. These values are calculated from the normalized cumulative sum of eigenvalues.}
	\label{tab:pca}
	\centering
	\begin{tabular}{ll}
		\toprule
		Number of & Percentage of \\
		Included Components & Dataset Explained \\
		\midrule
		1&0.90115488 \\ %
		2&0.99739416 \\ %
		3&0.99865254 \\ %
		4&0.99973184 \\ %
		5&0.99989146 \\ %
		6&0.99996631 \\ %
		7&0.99999949 \\ %
		8&1.0        \\ %
		\bottomrule
	\end{tabular}
\end{table} 

\subsection{Network Architechture}
The mapping from rank-optimized component weights to a projection image is highly nonlinear as it includes both a diffeomorphic deformation and radiographic projection.
We aim to approximate the inverse of this mapping with a deep convolutional neural network as a multi-output continuous regression problem.
A promising network architecture for this application is DenseNet \cite{Huang2017}. 
DenseNet has achieved state-of-the-art performance on several datasets while using fewer parameters and decreasing over-fitting. 
The direct connections between constant spatial layers allow for reusing features learned from previous layers and improve the flow of gradients throughout the network. 
Additionally, DenseNet is ideal for real-time applications due to reduced model complexity and depth.    

\begin{figure}[htbp]
	\centering
	\input{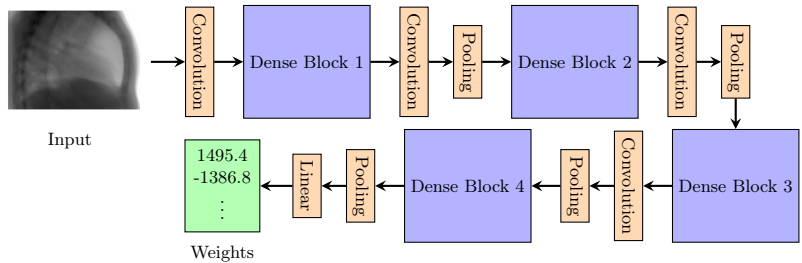}
	\caption{\label{fig:network} DenseNet architecture used to learn the inverse deformation-projection estimate. Each dense block represents a group of 4 feature layers and growth rate 4 with an included transition layer. A final linear layer recovers the weights for deformation components.}
\end{figure}

We use an efficient DenseNet model in PyTorch tailored to our application so that single-channel images are used as input \cite{Pleiss2017,paszke2017automatic}. 
The rank-optimized component weights used to produce the projection image are the regression points (Fig. \ref{fig:network}).
Summarized here for completeness, the DenseNet architecture convolves each input with a kernel size of 3 and filter number of $2\times k$ before being input to the first dense block, where $k$ is the growth rate. 
Each subsequent dense block consists of 8 convolutional layers all with filter size of 3 and a growth rate of 4. 
The 4 convolutional layers within a dense block are each preceded by a batch normalization layer and a rectified linear unit (ReLU) layer. 
The output from a dense block enters a transition block where batch normalization and non-linear activation are followed by a convolutional feature map reduction layer (compression=0.5) and spatial reduction factor of 2 in each dimension using max-pooling. 
A total of 4 dense and transition blocks are used to reduce the spatial dimensions to final feature maps.
The final layer of the network was a linear layer that regresses the feature maps to the rank-optimized component weights. 

\section{Results}
For our experiments, we use a 10-phase RCCT of a lung radiotherapy patient as the source data. 
Rank-optimized deformations were calculated as described by Foote \etal and summarized in section \ref{sec:rankdef} using the full-exhale (Phase 5) CT volume as a reference.
Using PCA, rank-optimized components of these deformations were determined and used to generate a DRR training dataset as described in \ref{sec:synth}.
In section \ref{ssec:train} we outline the specific training procedure for the neural network. 
Then in section \ref{ssec:spline} we test the neural network against unseen data from a breathing model.
Finally, section \ref{ssec:ct} describes generalization of the trained model on the phases of the RCCT dataset that were not used in training dataset generation.

\subsection{Training and Test Performance}
\label{ssec:train}
The network is trained for 300 epochs using a $L_1$ loss function with a learning rate of 0.1 and batch size of 2048.
The $L_1$ loss independently penalizes each deformation component which avoids preferential fitting of the larger-weighted component at any point.
The learning rate is decreased by a factor of 5 when the training loss plateaus within a relative threshold of $10^{-4}$ for the last 20 epochs. 
The dataset is randomly split 80-20 between training and on-line testing to monitor for over-fitting during training.
Training for 300 epochs completes in under 2 hours on a single Nvidia Quadro GV100 GPU.

\subsection{Spline Model Deformation Validation}
\label{ssec:spline}
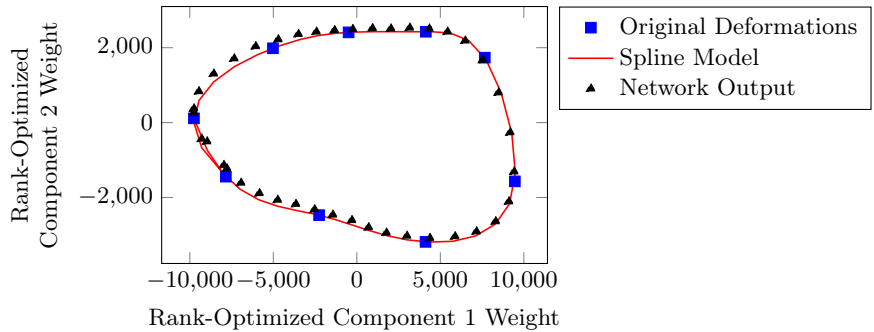
\begin{figure}[htb]
	\centering
	\begin{tikzpicture}%
\begin{axis}[xlabel=Rank-Optimized Component 1 Weight, ylabel style={align=center}, ylabel=Rank-Optimized\\Component 2 Weight, height=5cm, width=0.55\textwidth, legend style={cells={anchor=west}, legend pos=outer north east}, tick label style={font=\small},,scaled x ticks = false]
\addplot [mark=square*, only marks, color=blue] table [x=pc1, y=pc2, col sep=comma] {./figures/originalpcs.csv};
\addlegendentry[]{Original Deformations};

\addplot [no markers, semithick, color=red] table [x=spline_x, y=spline_y,col sep=comma] {figures/splinemodel.csv};
\addlegendentry[]{Spline Model};

\addplot [mark=triangle*, only marks, color=black, dashed] table [x=rec1, y=rec2, col sep=comma] {figures/synth_predicted_weights.csv};
\addlegendentry[]{Network Output};

\end{axis}
\end{tikzpicture}
	\caption{\label{fig:splinemodel} Deformation component weights for each of the 9 original components (blue squares) are control points for the spline-based breathing model (red line). DRR images derived from the weights along this spline model are input to the network. Resulting inferred weights (black triangles) closely align with the model.}
\end{figure}

A motion-subspace-based breathing model was created using a spline interpolation of the first two rank-optimized component weights from the 9 rank-optimized deformations (Fig. \ref{fig:splinemodel}). 
At any point along the model curve, the coordinate weights provide a linear combination of the rank-optimized component deformations to apply to the reference full-exhale volume to obtain a model volume for the corresponding breathing phase.
DRR projections through these model volumes in the same manner as the training dataset generation produced fluoroscopic images for evaluation. 
As with the training images, these projection images are preprocessed with variance equalization, normalization, histogram equalization, and $2\times$ downsampling.
Evaluation of these images through the trained network recovers the weights of the rank-optimized components.
Inference on a single Nvidia Quadro GV100 GPU has a throughput of 1113 images/second with the same PyTorch implementation -- significantly faster than real-time.
The relation of these weights to the original spline model is shown in Figure \ref{fig:splinemodel}.

Accuracy of the deformation recovered from the inferred weights is measured by maximum deformation distance error compared to the reconstruction of these deformations with the two rank-optimized deformation component weights directly from the model. 
From the 40 model points, the maximum error between the applied deformation and the recovered deformation was 1.22 mm. 

\subsection{RCCT Phase Patient Data and Geometric Validation}
\label{ssec:ct}
Until this point, DRR images derived from 9 of the 10 original 4DCT volumes have not been used as input for the network for either training or previous testing.
Rather than a synthetic spline model of respiratory phases, we now use the original CT volumes captured at 9 stages of the respiratory cycle.
These CT volumes inherently contain deformations that have not been included in any training or evaluation to this point, as the rank-constrained motion estimation produces imperfect deformations to these CT images.
No deformation is applied to these CT volumes as each volume represents this intrinsic respiratory deformation relative to the full-exhale anatomical state.
Evaluation of our framework on these intrinsic deformations effectively projects the true deformation into our 2-dimensional motion subspace.

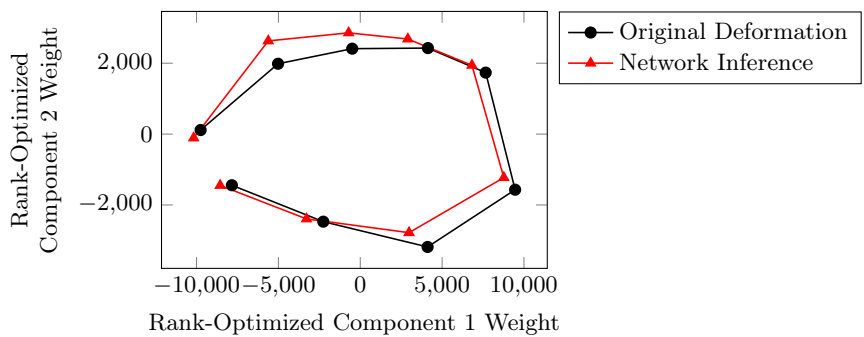
\begin{figure}[htbp]
	\centering
	\begin{tikzpicture}
\begin{axis}[xlabel=Rank-Optimized Component 1 Weight, ylabel style={align=center}, ylabel=Rank-Optimized\\Component 2 Weight,  height=5cm, width=0.55\textwidth, legend style={cells={anchor=west}, legend pos=outer north east}, tick label style={font=\small},scaled x ticks = false]

\addplot [mark=*, semithick, color=black] table [x=pc1, y=pc2, col sep=comma] {figures/originalpcs.csv};
\addlegendentry[]{Original Deformation};

\addplot [mark=triangle*, semithick, color=red] table [x=rec1, y=rec2, col sep=comma] {figures/patient_data_weights.csv};
\addlegendentry[]{Network Inference};

\end{axis}
\end{tikzpicture}
	\caption{\label{fig:realValidation} Weights of rank-optimized deformation components recovered by network on real patient CT data (red triangles) align well with original deformation weights for rank-optimized components (black circles). The respiratory cycle proceeds clockwise around the loop.}
\end{figure}

Exactly as in training dataset generation, DRR images through these (undeformed) volumes are calculated and preprocessed with variance equalization, normalization, histogram equalization, and downsampling. 
Evaluation of the network gives rank-optimized deformation component weights describing the intrinsic deformations that can be compared in the subspace of motion component weights with the weights directly recovered from original rank-constrained density motion estimation (Fig. \ref{fig:realValidation}).

\begin{figure}[htb]
	\centering
	\input{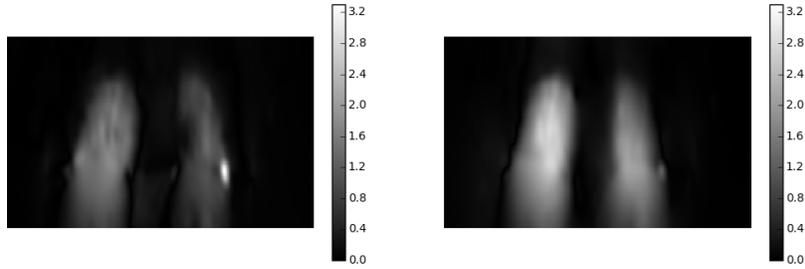}
	\caption{\label{fig:error} Slice of error map of calculated deformation error recovered by the deep learning framework. Phase 0 (left) and 9 (right) are selected as representative examples. Both general and localized errors are similar to the axial resolution of the CT scan (3mm).}
\end{figure}

These weights produced by the network are used to reconstruct a deformation field as a linear combination of the rank-optimized deformation component fields.
We calculate and report the error as a difference from the original rank-constrained motion estimation deformation distance that is recovered by both our deep learning framework in Figure \ref{fig:error}. 
Errors for the deformation recovered at each CT phase are also calculated against the deformations from rank-constrained motion estimation (Table \ref{tab:error}).
\begin{table}[htb]
	\caption{Deformation distance error for each phase's recovered deformation.}%
	\label{tab:error}
	\centering
	\begin{tabular}{lll}
		\toprule
		Phase \hspace*{1cm} & Average Distance Error (mm) \hspace*{0.2cm} &  Maximum Distance Error (mm)\\
		\midrule
		0&0.136&3.92 \\ %
		1&0.212&3.15 \\ %
		2&0.132&2.58 \\ %
		3&0.291&8.98 \\ %
		4&0.223&6.62 \\ %
		6&0.201&4.15 \\ %
		7&0.286&6.96 \\ %
		8&0.252&9.55       \\ %
		9&0.183&5.07   \\
		\bottomrule
	\end{tabular}
\end{table} 
\FloatBarrier

\section{Discussion}
In this paper, we have shown that estimation of anatomic position in the presence of breathing variability is possible with the combination of (1) rank-constrained density motion estimation for determination of motion components and (2) deep learning for subsequent identification of the weights of motion components in real-time.
While this framework is extensible to dimensions higher than 2, the rank-constrained nature of the deformation fields produces accurate results using only two dimensions.
This level of accuracy is a trade-off against the curse of dimensionality in both the dataset size and computational cost of training; however, using rank-constrained deformation components increases the accuracy of resulting deformations with the same number of dimensions.

The speed and accuracy attained by this framework is suitable for inclusion as a tumor position monitoring component of a conformal radiation therapy system.
Determination of 3D tumor location from noninvasive 2D radiographic projections via deep learning instead of variational optimization approaches to the 2D-3D deformation determination problem provides real-time results for conformal radiation therapy for lung tumors.

\section*{Acknowledgements}
This work was partially supported through research funding from the National Institute of Health (R01 CA169102 and R03 EB026132) and the Huntsman Cancer Institute. The authors are grateful for the support of NVIDIA Corporation by providing the GPU used for this research.
\bibliographystyle{splncs04}
\bibliography{ipmi2019_paper85}

\end{document}